# Integration of Text and Graph-based Features for Detecting Mental Health Disorders from Voice

Nasser Ghadiri [1] Rasoul Samani [1] Fahime Shahrokh [1]

## Abstract

With the availability of voice-enabled devices such as smart phones, mental health disorders could be detected and treated earlier, particularly post-pandemic. The current methods involve extracting features directly from audio signals. In this paper, two methods are used to enrich voice analysis for depression detection: graph transformation of voice signals, and natural language processing of the transcript based on representational learning, fused together to produce final class labels. The results of experiments with the DAIC-WOZ dataset suggest that integration of text-based voice classification and learning from low level and graph-based voice signal features can improve the detection of mental disorders like depression.

## 1. Introduction

Large-scale outbreaks such as Ebola virus disease (EVD) epidemic, Corona virus disease (Covid-19) pandemic, and other rapidly spreading diseases have a potential impact on mental health problems(Amsalem et al., 2021). Mental disorders like anxiety and depression have a negative impact on public heath efforts and require more efficient methods for detection to enable earlier treatment.

Computational methods for automatic detection of depression convey different types of input modalities, including user generated text on social media (Peng et al., 2019), interview transcripts (Gong & Poellabauer, 2017; Sun et al., 2017), and the voice signal of the interviewee (Toto et al., 2020; Dubagunta et al., 2019).

Recently, multi-modal approaches that integrate text and voice features have gained attention due to more accurate results for for depression detection (Alhanai et al., 2018; Lam et al., 2019; Ye et al., 2021).

Signal processing methods are commonly used for extracting features from voice. However, converting the audio signal into a graph provides a new perspective for the expert to analyse the human voice with a more enriched representation (Yela et al., 2019).

In this paper, we propose a novel method for multi-modal detection of depression from text and voice, that employs lowlevel audio features, graph-based feature extraction from the audio signal and the transcript as inputs. The major contribution of this work are (i) a novel method of depression detection from text using pretrained models and (ii) mapping of voice signal to a network for extracting graph-based features in depression detection. We evaluated our method on a standard DAIC-WOZ depression detection benchmark and showed that it ties with existing single-modal audiobased detection, and outperforms multi-modal methods with %82.4 F1-score and %86.6 accuracy.

The rest of the paper is organised as follows. Section 2 gives an overview of related work. Section 3 elaborates on the proposed methods The experimental results are presented in Section 4, and Section 5 concludes the paper.

## 2. Related Work

In this section, the related work is reviewed in three categories including text-based, voice-based and multi-modal method for depression detection.

[1] Department of Electrical and Computer Engineering, Isfahan University of Technology, Isfahan, Iran. Correspondence to: Nasser Ghadiri <nghadiri@iut.ac.ir>.

## 2.1. Text-based Depression Detection

The text extracted from the human speech, and validated by the experts as transcripts is used for depression detection.
Shared tasks such as 2017 Audio/Visual Emotion Challenge (AVEC) advocate the text processing methods. Two notable works are (Gong & Poellabauer, 2017) that builds a topic model for the analysis of the interview recording. In (Sun et al., 2017) a random forest method with a selected features is constructed according to the transcript in different levels of depression.

User generated text from social media are also used for depression detection including (Peng et al., 2019). A comprehensive survey for this category is provided by (Skaik & Inkpen, 2021). For this type of input, the ground-truth is usually unknown or is known with low confidence. The transformer-based methods with pretrained models have provided higher accuracy than classic methods (Lin et al., 2020).

## 2.2. Voice-based Depression Detection

Human speech signal is affected by neurophysiological changes that might occur during depression, so analysing voice signals helps in finding out whether the disease is present. In (Ma et al., 2016) the authors propose two deep learning architecture named the DepAudioNet that uses Convolutional Neural Network (CNN) and Long Short-Term Memory (LSTM) with a random sampling strategy for the training phase that handles imbalanced data. A CNN architecture is also exploited in (Dubagunta et al., 2019) with applying different types of filters to input signal. The output is the probability of depression.

The authors in (Tlachac et al., 2020) focused on audio feature engineering and machine learning models based on built on Betti curves that provides higher accuracy than compared voice-based methods. The novelty of (Toto et al., 2020) is in the signal processing phase with a Sliding Window Sub-clip Pooling (SWUP) method.

Using voice features as biomarkers was studied by (Zhang et al., 2020). They collected voice samples from the Mental Health America (MHA) website and extracted acoustic, prosodic and linguistic features for predicting depression and suicidality. The authors in (Bailey & Plumbley, 2020) tried to removed gender bias through balancing the training audio data. In (Shin et al., 2021) specific features of voice were used for depression detection including pitch and the ratio of the actual utterance to the utterance time.

## 2.3. Multi-modal Voice and Text Approaches

Due to the limited accuracy of single-modal works, multimodal architectures have gained attention recently. Audio, video and semantic features are fused by (Williamson et al., 2016) resulting in increased accuracy compared to single modalities. An data-driven model for integrating voice and text is proposed by (Alhanai et al., 2018). The model uses logistics regression and a LSTM deep neural networks architecture for each modality (audio and text) that are trained separately. A multi-modal model is also trained by integrating audio and text into feedforward layers. The results show that using LSTM for sequence modeling provides higher accuracy and robustness for depression detection, compared to context-free modeling that gives more discriminating power.

In (Lam et al., 2019) a context-aware multi-modal model uses data augmentation procedure based on topic modelling. By exploiting transformer architecture and deep 1D Convolutional Neural Network (CNN) the audio signal features are modeled to achieve higher performance for audio and text modalities. Combining the modalities has also resulted in higher F1 scores compared to other depression detection systems.

A specific data gathering method from Chinese speaking persons was performed by (Ye et al., 2021). Low level audio features extracted from voice, and textual features extracted from the answers to a set of question, and fused together using deep neural networks. A hybrid of deep learning, SVM, and random forests also proposed by (Yang et al., 2021) that uses text, voice and video to detect depression.

To the best knowledge of authors, no existing multi-modal method uses graph-based features of audio signals for depression detection. We will elaborate our model on this novel set of features, as well as state-of-the-art pretrained models for handling text.

## 3. Proposed Method

In Section 3.1, we will present the concept of a visibility graph as an alternate approach for analysing signals and time-series including audio signals. The proposed multimodal method for depression detection will be described in Section 3.2.

## 3.1. Mapping of Signals to Graphs

Mapping from a time series to a complex network was proposed by (Lacasa et al., 2008). For this purpose, a visibility graph (VG) maps an input time-series like an audio signal into a complex network. Two given data points $(t_a, y_a)$ and $(t_b, y_b)$ are visible to each other if any other data point $(t_c, y_c)$ that is placed between them satisfies the following condition shown by Eq. (1):

$$y_c < y_b + (y_a - y_b)\frac{(t_b - t_c)}{(t_b - t_a)}, \qquad (1)$$

The two points that satisfy the visibility condition will be mapped into two nodes in the visibility graph, and an edge connects them. The process of mapping a signal into the constructed visibility graph is shown in Fig. 1. The peaks of the input audio signal are detected through the envelope, and placed as graph nodes. Every peak is then connected to all nodes that are visible without obstacles, yielding the edges of the graph. The resulting visibility graph is also shown.

The visibility graph potentially contributes to enriched understanding of the signal in two ways. First, the topology and visual aspects of the graph might help the expert to gaining a deeper insight into the underlying our multi-modal depression detection that will be described in the next subsection.

3.2. Overall Process

We propose a hybrid architecture for multi-modal depression detection from text and voice inputs. The overall process is shown in Fig. 2. The model consists of two main pipelines for processing text and voice signal. The text pipeline gets the transcription as input and passes it through a transformerbased deep learning for sentence classification. The voice pipeline comprises three main parts that extract different types of features from the patient voice and feed them into the prediction model that classifies the voice signal. At the final stage, the outputs from all machine learning models are fused using a trained random forest classifier and the final label of the patient is predicted. Each part of the proposed method are presented in the following. The details of dataset and specific preprocessings will be described in Section 4.1 and 4.2 respectively.

*Text model* As mentioned in Section 2, deep learning and signal. For this purpose, graph algorithm such as community detection are commonly used. This aspect is applicable to smaller graphs where visual inspection is feasible.

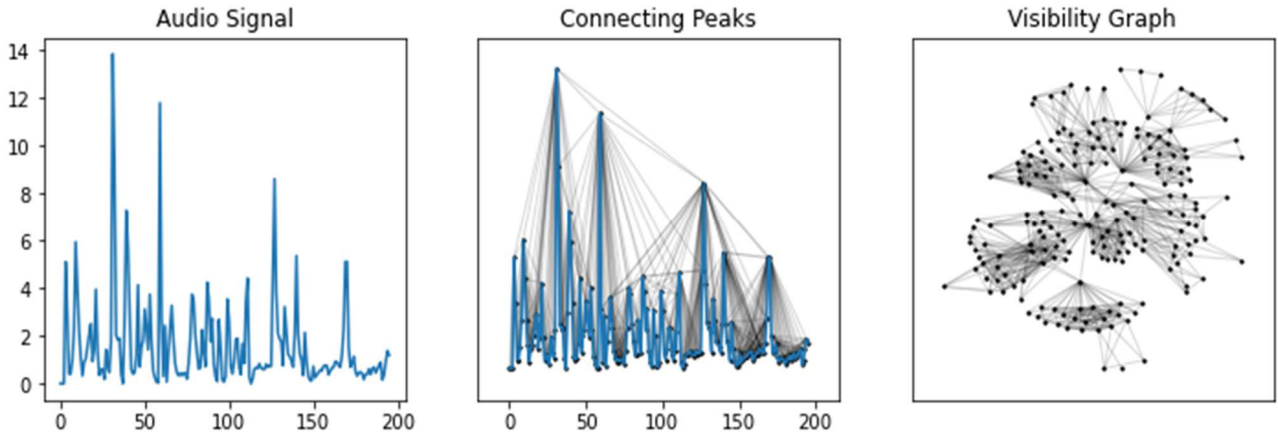

*Figure 1.* The visibility graph

Secondly, the network features such as node degrees, centrality and many other features could be extracted from the constructed network, and processed by a machine-learning model. Different domains such as vibration analysis use the second aspect (Zhang et al., 2018). In (Yela et al., 2019) presents a single-modal approach to create a graph for comparing the audio signals and computing the harmonic-based similarity. We will also exploit the second aspect of the visibility graph as part of specifically, transformer-based architectures such as BERT provide higher accuracy than classic methods for text classification. We also use transformer architecture in our proposed method for depression detection. The preprocessed transcription question and answer text is given to the BERT model with the patient label. The BERT model is then fine tuned on the Sequence Classification task for tuning the binary classifier. After fine-tuning the model, the Scalable Depression Prediction voting mechanism is exploited that performs according to the predicted labels.

*Voice model* The voice input to our model passes through three different feature extraction modules. In the MFCC

module, low level features are extracted from the input signal that include MFCC, Spectrogram, and Mel-Spectrogram. In the openSMILE module (Eyben et al., 2010), we extracts 88 features based on the eGeMAPSv02 feature set. The third module extract graph-based features from the input voice and will be described in the following section.

*Graph-based voice features* One of the key contributions of our proposed model are the graph-based features of the speech signal. As noted in Section 3.1, the visibility graph is constructed from the voice input as a time-series. In addition to visually inspection of the graph that could help the experts in detecting depression, different types of graph features could be extracted to train a machine learning model. Our model employs eight graph features for this purpose that will be described in Section 4.

*Fusion* We first compute the average of three different voice-based detection modules, and merge the result with the output from the text-processing module to determine the final class.

### 3.3. Scalable Depression Prediction

There are multiple question and answer sections for every patient in the dataset. Our model predicts the depression label based on every single pair of question and answer for each patient. However, we also need to integrate all single predictions to predict the overall label for any patient. For this purpose, we use the method proposed by (Huang et al., 2019) shown by Eq. (2) to compute the probability of depression for a given patient, based on predictions by n sequences of text:

$$P(depr = 1|h_{patient}) = \frac{P_{max}^n + P_{max}^n n/c}{1 + n/c} \quad (2)$$

The scaling factor *c* controls the influence of the number of subsequences n, and $h_{patient}$) is represents all of a patient's texts. The maximum and mean probabilities of depression over n subsequences are $P_{max}^n$ and $P_{mean}^n$.

## 4. Experiments

In this section we present the dataset, our experiments based on proposed model, and discuss the results. For our experiments we used Google Colab Pro with 25GB of RAM and Tesla K80 GPU using 12GB of memory. Other settings and parameters are as follows. The BERT-Version was BERTbase, MaxLength=512, BatchSize=8,16, LearningRate was set to 2e-5,3e-5, 5e-5 and the number or epochs was 10.

### 4.1. Dataset

The Distress Analysis Interview Corpus/Wizard-of-Oz set (DAIC-WOZ) dataset (Gratch et al., 2014; DeVault et al., 2014) comprises voice and text samples from 189 interviewed healthy and control persons, as wells as their PHQ-8 depression detection questionnaire. This dataset is commonly used in many of the depression detection research works, including (Gong & Poellabauer, 2017; Sun et al., 2017) for text-based detection, (Dubagunta et al., 2019; Toto et al., 2020; Tlachac et al., 2020) for voice-based detection, and in multi-modal architectures such as (Alhanai et al., 2018; Yang et al., 2021). We also used both voice and textual data (interview transcripts) extracted from this dataset for our experiments.

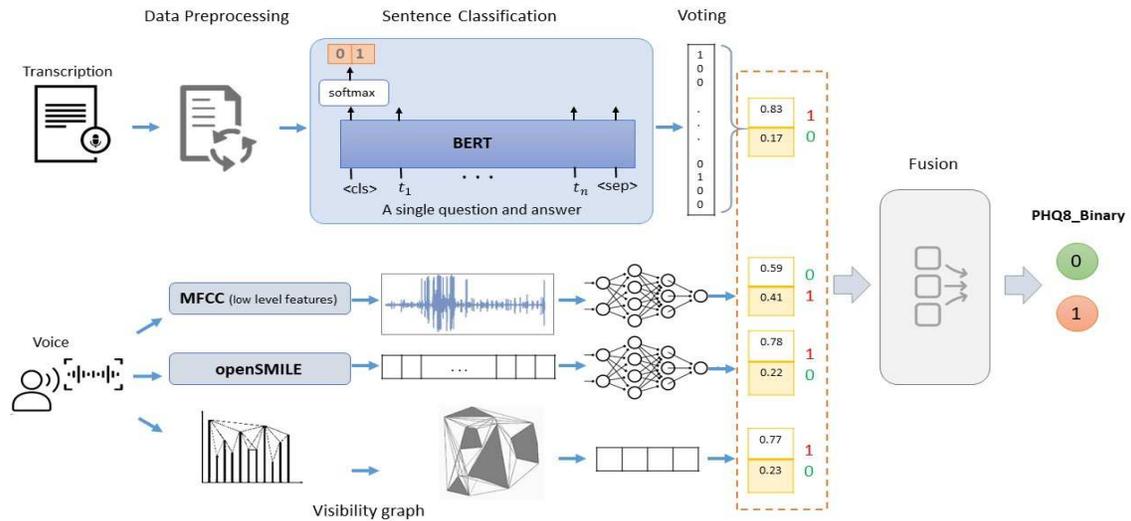

*Figure 2.* The overall process

### 4.2. Preprocessing

The dataset contains a total number of 189 interviews. Nine interviews (318, 321, 341, 362, 373, 444, 451, 458, 480) are exluded due problems such as text and voice being out of sync, extra noise, and intermittent microphone. The first and last sentences such as "Okay I think I've asked everything I need to" and "Hi I'm Ellie thanks for coming in today" were deleted. Punctuations and signs such as parenthesis and SYNC symbols removed. Abbreviations like "feel lately" (how have you been feeling lately) for which full expressions exist also deleted. Finally, the transcript text for each patient separated into question and answer sections.

Multiple answers and a single question are concatenated into a paragraph. This is a key difference between our work and other methods that we also process the question text. A sample is shown in Figure 3. The size of training, test and validation sets were 2082, 1022 and 622 respectively.

| Person | Text |
|---|---|
| Ellie | do you consider yourself more shy or outgoing |
| Participant | i'm definitely more shy |
| Participant | but i can be outgoing when i'm comfortable with people but |
| Participant | initially i'm pretty shy |

*Figure 3.* Sample question and multiple answers that are merged in preprocessing.

*Table 1.* Text-based classification results for pretrained models.

| Model | Acc. | Prec. | Recall | F1 |
|---|---|---|---|---|
| BERT-UNCASED | 0.82 | 0.68 | 0.78 | 0.72 |
| BERT-CASED | 0.75 | 0.60 | 0.64 | 0.62 |
| ALBERT | 0.63 | 0.46 | 0.49 | 0.47 |
| ROBERTA | 0.77 | 0.66 | 0.42 | 0.52 |
| CLINICALBERT | 0.73 | 0.54 | 0.75 | 0.63 |
| DISTILBERT | 0.75 | 0.61 | 0.57 | 0.59 |
| BLUEBERT | 0.77 | 0.66 | 0.57 | 0.61 |

### 4.3. Text-based detection

We fine tuned different pretrained models to detect depression from the transcripts. The results are reported in Table 1. The best F1-scores are achieved by the general BERT-baseuncased model. Although other specific BERT models were also tuned, but the results suggest that due to the general, non-specific words used by the patients in the interview, the base BERT model stands well above specific BERT models.

The Receiver Operating Characteristic (ROC) curve is shown in Figure 4. Based on the form of the ROC curve and the Area Under the Curve (AUC) value of 88 percent, the proposed text processing method could effectively detect depressed patients as True Positives.

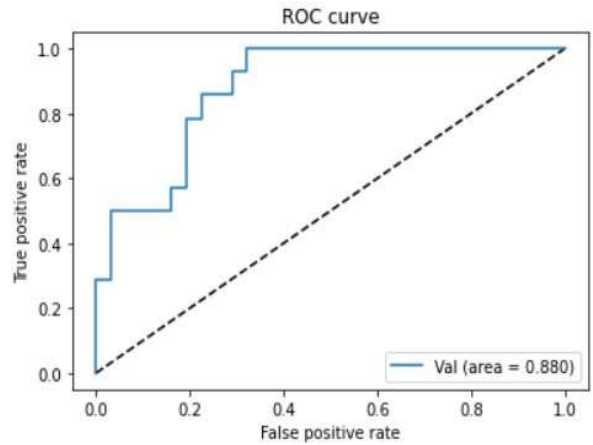

*Figure 4.* The ROC curve for text-based classification.

### 4.4. Voice-based detection

For the Visibility Graph (VG) part of our model as described in Section 3.1, we extracted eight features from the resulting network including average degree, average clustering coefficient, density, transitivity, diameter, local efficiency, global efficiency, and average shortest path. This enriched graph-based features complement our proposed multi-modal depression detection architecture. The results are shown in Figure 5. It shows that the graph generated features contribute to discriminating depression and non-depression classes.

The single-modal experiments results and their fusion are reported in Table 2. It can be observed that in single-modal detection, our text-based method has achieved %72.7 F1score that is higher than other single-modal scores. The fusion of all prediction models based on voice and text has resulted in %82.4 F1-score that outperforms other models.

## 5. Conclusion

In this paper we proposed a hybrid architecture for the prediction of depression from two modalities of voice and text. The transformer-based deep learning architecture with fine tuning of pretrained models achieved higher accuracy than existing methods. Moreover, our novel graph-based features extracted through the mapping of speech signal into a complex network contributed to even more accurate results. Future work includes employing graph-embedding, adding other modalities such as video into the multi-modal architecture, and improved fusion methods.

## Accessibility TBD.

## Software and Data

The DAIC-WOZ dataset is available by getting permission from the dataset owner. The source code for this paper is available for review, and will be made publicly available upon the acceptance.

## Acknowledgements TBD.

## References

Alhanai, T., Ghassemi, M., and Glass, J. Detecting depression with audio/text sequence modeling of interviews. *Proc. Annu. Conf. Int. Speech Commun. Assoc. INTERSPEECH*, 2018-Septe(September):1716–1720, 2018. doi: 10.21437/Interspeech.2018-2522.

Amsalem, D., Dixon, L. B., and Neria, Y. The Coronavirus Disease 2019 (COVID-19) Outbreak and Mental Health: Current Risks and Recommended Actions, jan 2021. ISSN 2168622X.

Bailey, A. and Plumbley, M. D. Gender Bias in Depression Detection Using Audio Features. In *EUSIPCO 2021*. University of Surrey, 2020. ISBN 9789082797060. doi: 10.23919/EUSIPCO54536.2021.9615933.

DeVault, D., Artstein, R., Benn, G., Dey, T., Fast, E., Gainer, A., Georgila, K., Gratch, J., Hartholt, A., Lhommet, M., Lucas, G., Marsella, S., Morbini, F., Nazarian, A., Scherer, S., Stratou, G., Suri, A., Traum, D., Wood, R., Xu, Y., Rizzo, A., and Morency, L. P. SimSensei kiosk: A virtual human interviewer for healthcare decision support. In *13th Int. Conf. Auton. Agents Multiagent Syst. AAMAS 2014*, volume 2, pp. 1061–1068, 2014. ISBN 9781634391313.

Dubagunta, S. P., Vlasenko, B., and Magimai-Doss, M. Learning Voice Source Related Information for Depression Detection. In *ICASSP, IEEE Int. Conf. Acoust. Speech Signal Process. - Proc.*, volume 2019-May, pp. 6525–6529. Institute of Electrical and Electronics Engineers Inc., may 2019. doi: 10.1109/ICASSP.2019.8683498.

Eyben, F., Wollmer, M., and Schuller, B. OpenSMILE - ¨ The Munich versatile and fast open-source audio feature extractor. In *MM'10 - Proc. ACM Multimed. 2010 Int. Conf.*, pp. 1459–1462, 2010. ISBN 9781605589336. doi: 10.1145/1873951.1874246.

Gong, Y. and Poellabauer, C. Topic modeling based multimodal depression detection. In *AVEC 2017 - Proc. 7th Annu. Work. Audio/Visual Emot. Challenge, co-located with MM 2017*, pp. 69–76. Association for Computing Machinery, Inc, oct 2017. ISBN 9781450355025. doi: 10.1145/3133944.3133945.

Gratch, J., Artstein, R., Lucas, G., Stratou, G., Scherer, S., Nazarian, A., Wood, R., Boberg, J., De Vault, D., Marsella, S., Traum, D., Rizzo, S., and Morency, L. P. The distress analysis interview corpus of human and computer interviews. In *Proc. 9th Int. Conf. Lang. Resour. Eval. Lr. 2014*, pp. 3123–3128, 2014.

Table 2. Audio, text and graph-based classification results.

| MODEL | FEATURES | F1 | PREC. | RECALL | ACC. |
|---|---|---|---|---|---|
| (WILLIAMSON ET AL., 2016) | AUDIO | 50.0 | - | - | - |
| (WILLIAMSON ET AL., 2016) | FUSION | 81.0 | - | - | - |
| (TLACHAC ET AL., 2020) | AUDIO | 48.7 | - | - | 60.3 |
| (BAILEY & PLUMBLEY, 2020) | AUDIO | 65.8 | - | - | - |
| (TOTO ET AL., 2020) | AUDIO | 63.1 | - | - | - |
| OUR WORK | TEXT | 72.7 | 68.8 | 78.6 | 82.2 |
| OUR WORK | AUDIO (LOW LEVEL) | 64.0 | 62.4 | 64.6 | - |
| OUR WORK | AUDIO (OPENSMILE) | 63.4 | 61.1 | 66.7 | 61.0 |
| OUR WORK | AUDIO (VG) | 59.0 | 56.2 | 62.1 | - |
| OUR WORK | FUSION | 82.4 | 70.0 | 100.0 | 86.6 |

Huang, K., Altosaar, J., and Ranganath, R. ClinicalBERT: Modeling Clinical Notes and Predicting Hospital Readmission. apr 2019. ISSN 2331-8422. URL http://arxiv.org/abs/1904.05342.

Lacasa, L., Luque, B., Ballesteros, F., Luque, J., and Nuno,˜ J. C. From time series to complex networks: The visibility graph. *Proc. Natl. Acad. Sci. U. S. A.*, 105(13):4972–4975, apr 2008. doi: 10.1073/pnas.0709247105.

Lam, G., Dongyan, H., and Lin, W. Context-aware Deep Learning for Multi-modal Depression Detection. In *ICASSP, IEEE Int. Conf. Acoust. Speech Signal Process. - Proc.*, volume 2019-May, pp. 3946–3950. Institute of Electrical and Electronics Engineers Inc., may 2019. doi: 10.1109/ICASSP.2019.8683027.

Lin, C., Hu, P., Su, H., Li, S., Mei, J., Zhou, J., and Leung, H. SenseMood: Depression detection on social media. In *ICMR 2020 - Proc. 2020 Int. Conf. Multimed. Retr.*, pp. 407–411. Association for Computing Machinery, Inc, jun 2020. ISBN 9781450370875. doi: 10.1145/3372278.3391932.

Ma, X., Yang, H., Chen, Q., Huang, D., and Wang, Y. DepAudioNet: An efficient deep model for audio based depression classification. In *AVEC 2016 - Proc. 6th Int. Work. Audio/Visual Emot. Challenge, co-located with ACM Multimed. 2016*, pp. 35–42. Association for Computing Machinery, Inc, oct 2016. ISBN 9781450345163. doi: 10.1145/2988257.2988267.

Peng, Z., Hu, Q., and Dang, J. Multi-kernel SVM based depression recognition using social media data. *Int. J. Mach. Learn. Cybern.*, 10(1):43–57, jun 2019. ISSN 1868808X. doi: 10.1007/s13042-017-0697-1.

Shin, D., Cho, W. I., Park, C. H. K., Rhee, S. J., Kim, M. J., Lee, H., Kim, N. S., and Ahn, Y. M. Detection of minor and major depression through voice as a biomarker using machine learning. *J. Clin. Med.*, 10(14):3046, jul 2021. ISSN 20770383. doi: 10.3390/jcm10143046.

Skaik, R. and Inkpen, D. Using Social Media for Mental Health Surveillance: A Review. *ACM Comput. Surv.*, 53 (6), dec 2021. ISSN 15577341. doi: 10.1145/3422824.

Sun, B., Yu, L., Zhang, Y., Xu, Q., Wang, Z., He, J., and Li, D. A random forest regression method with selectedtext feature for depression assessment. In *AVEC 2017 Proc. 7th Annu. Work. Audio/Visual Emot. Challenge, colocated with MM 2017*, pp. 61–68. Association for Computing Machinery, Inc, oct 2017. ISBN 9781450355025. doi: 10.1145/3133944.3133951.

Tlachac, M. L., Sargent, A., Toto, E., Paffenroth, R., and Rundensteiner, E. Topological Data Analysis to Engineer Features from Audio Signals for Depression Detection. In *Proc. - 19th IEEE Int. Conf. Mach. Learn. Appl. ICMLA 2020*, pp. 302–307. Institute of Electrical and Electronics Engineers Inc., dec 2020. doi: 10.1109/ICMLA51294.2020.00056.

Toto, E., Tlachac, M. L., Stevens, F. L., and Rundensteiner, E. A. Audio-based Depression Screening using Sliding Window Sub-clip Pooling. In *Proc. - 19th IEEE Int. Conf. Mach. Learn. Appl. ICMLA 2020*, pp. 791–796. Institute of Electrical and Electronics Engineers Inc., dec 2020. doi: 10.1109/ICMLA51294.2020.00129.

Williamson, J. R., Godoy, E., Cha, M., Schwarzentruber, A., Khorrami, P., Gwon, Y., Kung, H. T., Dagli, C., and Quatieri, T. F. Detecting depression using vocal, facial and semantic communication cues. In *AVEC 2016 - Proc. 6th Int. Work. Audio/Visual Emot. Challenge, colocated with ACM Multimed. 2016*, pp. 11–18. Association for Computing Machinery, Inc, oct 2016. ISBN 9781450345163. doi: 10.1145/2988257.2988263.

Yang, L., Jiang, D., and Sahli, H. Integrating Deep and Shallow Models for Multi-Modal Depression AnalysisHybrid Architectures. *IEEE Trans. Affect. Comput.*, 12(1): 239–253, jan 2021. doi: 10.1109/TAFFC.2018.2870398.

Ye, J., Yu, Y., Wang, Q., Li, W., Liang, H., Zheng, Y., and Fu, G. Multi-modal depression detection based on emotional audio and evaluation text. *J. Affect. Disord.*, 295(February):904–913, 2021. doi: 10.1016/j.jad.2021.08.090.

Yela, D. F., Stowell, D., and Sandler, M. Spectral visibility graphs: Application to similarity of harmonic signals. *Eur. Signal Process. Conf.*, 2019-Septe, 2019. ISSN 22195491. doi: 10.23919/EUSIPCO.2019.8903056.

Zhang, L., Duvvuri, R., Chandra, K. K., Nguyen, T., and Ghomi, R. H. Automated voice biomarkers for depression symptoms using an online cross-sectional data collection initiative. *Depress. Anxiety*, 37(7):657–669, jul 2020. doi: 10.1002/da.23020.

Zhang, Z., Qin, Y., Jia, L., and Chen, X. Visibility graph feature model of vibration signals: A novel bearing fault

diagnosis approach. *Materials (Basel).*, 11(11):2262, nov 2018. ISSN 19961944. doi: 10.3390/ma11112262.

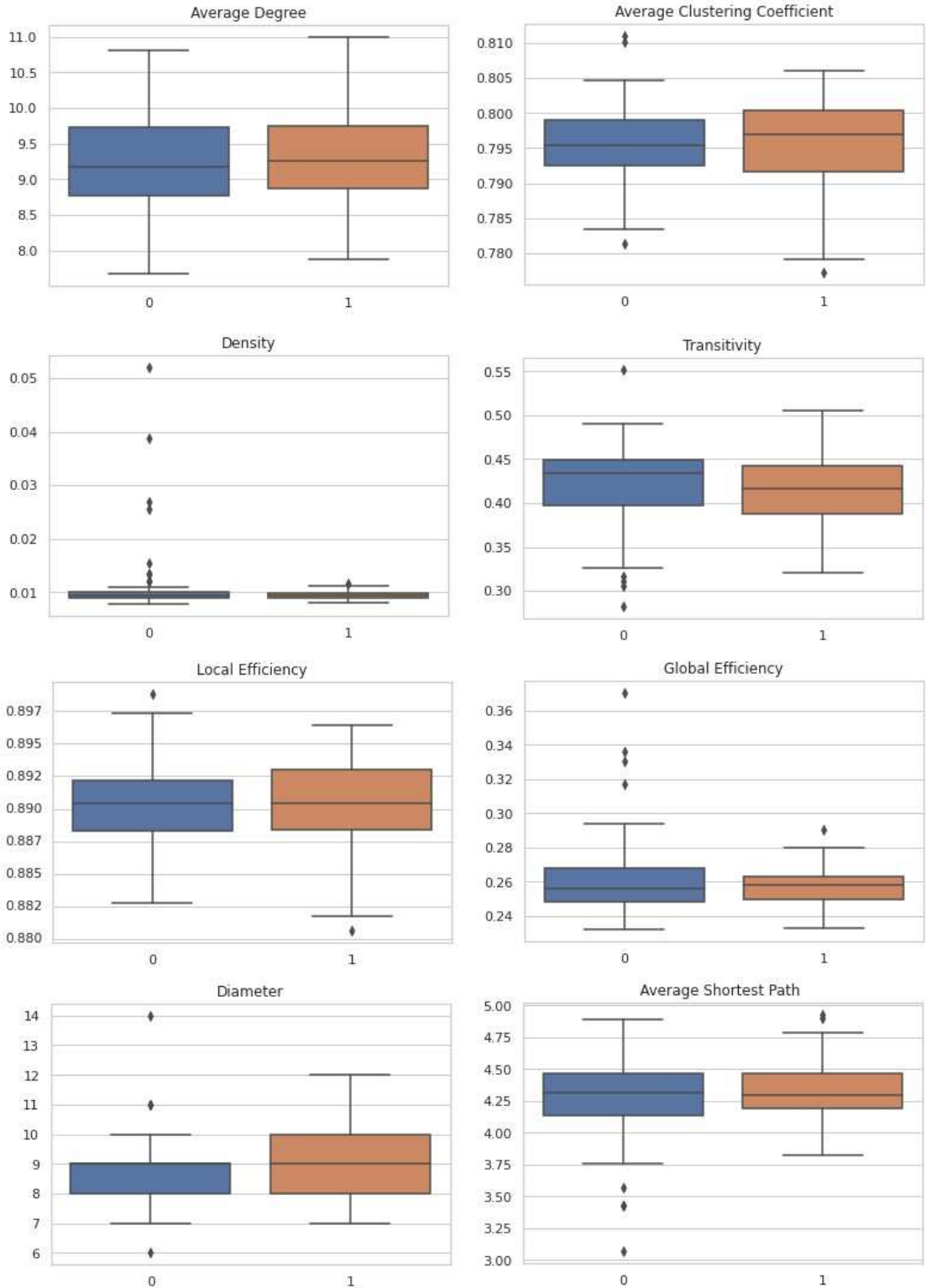

*Figure 5.* Analysis of the VG features.